%% file: sample-sigconf.tex
\pdfoutput=1
\documentclass[sigconf]{acmart}
\usepackage{graphicx} 
\usepackage{float}
\usepackage{booktabs} 
\usepackage{enumitem,kantlipsum}
\usepackage{threeparttable}
\usepackage{algorithm}
\usepackage{algpseudocode}
\usepackage{multirow}
\usepackage{array}
\usepackage{booktabs}
\usepackage{subfig}
\usepackage{pgfplots}
\usepackage{tikz}
\usepgfplotslibrary{groupplots}

\copyrightyear{2020} 
\acmYear{2020} 
\setcopyright{acmcopyright}\acmConference[SIGIR '20]{Proceedings of the 43rd International ACM SIGIR Conference on Research and Development in Information Retrieval}{July 25--30, 2020}{Virtual Event, China}
\acmBooktitle{Proceedings of the 43rd International ACM SIGIR Conference on Research and Development in Information Retrieval (SIGIR '20), July 25--30, 2020, Virtual Event, China}
\acmPrice{15.00}
\acmDOI{10.1145/3397271.3401147}
\acmISBN{978-1-4503-8016-4/20/07}

\begin{document}
\fancyhead{}

\title{Self-Supervised Reinforcement Learning for \\Recommender Systems}
\author{Xin Xin}
\authornote{Part of this work is done when taking an internship in Telefonica Research, Barcelona.}
\affiliation{%
  \institution{University of Glasgow}
}
\email{x.xin.1@research.gla.ac.uk}

\author{Alexandros Karatzoglou}
\affiliation{%
  \institution{Google Research, London}
}
\email{alexkz@google.com}

\author{Ioannis Arapakis}
\affiliation{%
  \institution{Telefonica Research, Barcelona}
}
\email{arapakis.ioannis@gmail.com}

\author{Joemon M. Jose}
\affiliation{%
  \institution{University of Glasgow}
}
\email{Joemon.Jose@glasgow.ac.uk}





\input{abstract.tex}
%
%


\maketitle

\input{introduction}
\input{preliminaries.tex}
\input{method.tex}
\input{experiment.tex}
\input{conclusion.tex}

\bibliographystyle{ACM-Reference-Format}
\bibliography{sample-bibliography}

\end{document}

%% file: abstract.tex
\begin{abstract}
In session-based or sequential recommendation, it is important to consider a number of factors like long-term user engagement, multiple types of user-item interactions such as clicks, purchases etc. The current state-of-the-art supervised approaches fail to model them appropriately. Casting sequential recommendation task as a reinforcement learning (RL) problem is a promising direction. A major component of RL approaches is to train the agent through interactions with the environment. However,  it is often problematic to train a recommender in an on-line fashion due to the requirement to expose users to irrelevant recommendations. As a result, learning the policy from logged implicit feedback is of vital importance, which is challenging due to the \emph{pure off-policy setting} and \emph{lack of negative rewards (feedback)}. 

In this paper, we propose \emph{self-supervised reinforcement learning} for sequential recommendation tasks. Our approach augments standard recommendation models with two output layers: one for self-supervised learning and the other for RL. The RL part acts as a regularizer to drive the supervised layer focusing on specific rewards (e.g., recommending items which may lead to purchases rather than clicks) while the self-supervised layer with cross-entropy loss provides strong gradient signals for parameter updates. Based on such an approach, we propose two frameworks namely \emph{Self-Supervised Q-learning} (SQN) and \emph{Self-Supervised Actor-Critic} (SAC). We integrate the proposed frameworks with four state-of-the-art recommendation models. Experimental results on two real-world datasets demonstrate the effectiveness of our approach.
\end{abstract}

%% file: introduction.tex
\section{Introduction}
Generating next item recommendation from sequential user-item interactions in a session (e.g., views, clicks or purchases) is one of the most common use cases in domains of recommender systems, such as  e-commerce, video \cite{reinforce-e-commerce} and music recommendation \cite{nextitnet}. 
Session-based and sequential recommendation models have often been trained with {\it self-supervised learning}, in which the model is trained to predict the data itself instead of some ``external'' labels. For instance, in language modeling the task is often formulated as predicting the next word given the previous $m$ words. The same procedure can be utilized to predict the next item the user may be interested given past interactions, see e.g., \cite{gru4rec,nextitnet,SASRec}. However, this kind of approaches can lead to sub-optimal recommendations since the model is purely learned by a loss function defined on the discrepancy between model predictions and the self-supervision signal. Such a loss may not match the  expectations from the perspective of recommendation systems (e.g., long-term engagement). Moreover, there can be multiple types of user signals in one session, such as clicks, purchases  etc. How to leverage multiple types of user-item interactions to improve recommendation objectives (e.g., providing users recommendations that lead to real purchases) is also an important research question.

Reinforcement Learning (RL) has achieved impressive advances in game control \cite{alphogo,humanlevelcontrol} and related fields. A RL agent is trained to take actions given the state of the environment it operates in with the objective of getting the maximum long-term cumulative rewards. A recommender system aims (or should aim) to provide recommendations (actions) to users (environment) 
with the objective of maximising the long-term user satisfaction (reward) with the system. Since in principle the reward schema can be designed at will, RL allows to create models that can serve multiple objectives such as diversity and novelty. As a result, exploiting RL for recommendation has become a promising research direction. There are two classes of RL methods: model-free RL algorithms and model-based RL algorithms.

Model-free RL algorithms need to interact with the environment to observe the feedback and then optimize the policy. Doing this in an on-line fashion is typically unfeasible in commercial recommender systems since interactions with an under-trained policy would affect the user experience. A user may quickly abandon the service if the recommendations don't match her interests. A typical solution is learning off-policy from the logged implicit feedback. However, this poses the following challenges for applying RL-based methods:
\begin{enumerate}[leftmargin=*]
\item Pure off-policy settings. The policy should be learned from fixed logged data without interactions with the environment (users). Hence the data from which the RL agent is trained (i.e., logged data) will not match its policy. 
\cite{googlewsdmoffpolicycorrection} proposed to use propensity scores to perform off-policy correction but this kind of methods can suffer from unbounded high variances \cite{munos2016safeandefficient}. 
\item Lack of data and negative rewards. RL algorithms are data hungry, traditional techniques overcome this by either simulating the environments or by running RL iterations in controlled environments (e.g. games, robots). This is challenging in the case of recommendations especially considering the large number of potential actions (available items). Moreover, in most cases learning happens from implicit feedback. The agent only knows which items the user has interacted with but has no knowledge about what the user dislikes. In other words, simply regressing to the Bellman equation \cite{bellman1966dynamic} (i.e., Q-learning) wouldn't lead to good ranking performance when there is no negative feedback since the model will be biased towards positive relevance values. 
\end{enumerate}
An alternative to off-policy training for recommender systems is the use of model-based RL algorithms. In model-based RL, one first builds a model to simulate the environment. Then the agent can learn by interactions with the simulated environment \cite{chen2019generativeusermodel,shi2019virtual}. These two-stage methods heavily depend on the constructed simulator. Although related methods like generative adversarial networks (GANs) \cite{GAN} achieve good performance when generating images, simulating users' responses is a much more complex task.

In this paper, we propose \emph{self-supervised reinforcement learning} for recommender systems. The proposed approach serves as a learning mechanism and can be easily integrated with existing recommendation models. More precisely, given a sequential or session-based recommendation model, the (final) hidden state of this model can be seen as it's output as this hidden state is multiplied with the last layer 
to generate recommendations \cite{gru4rec,nextitnet,caser-rec,SASRec}. We augment these models with two final output layers (heads). One is the conventional self-supervised head\footnote{For simplicity, we use ``self-supervised'' and ``supervised'' interchangeable in this paper. Besides, ``head'' and ``layer'' are also interchangeable.} trained with cross-entropy loss to perform ranking while the second is trained with RL based on the defined rewards such as long-term user engagement, purchases, recommendation diversity and so on.
For the training of the RL head, we adopt double Q-learning which is more stable and robust in the off-policy setting \cite{double-q-learning}. The two heads complement each other during the learning process. The RL head serves as a regularizer to introduce desired properties to the recommendations while the ranking-based supervised head can provide 
negative signals for parameter updates. We propose two frameworks based on such an approach: (1) \emph{Self-Supervised Q-learning} (SQN) co-trains the two layers with a reply buffer generated from the logged implicit feedback; (2) \emph{Self-Supervised Actor-Critic} (SAC) treats the RL head as a critic measuring the value of actions in a given state while the supervised head as an actor to perform the final ranking among 
candidate items.
As a result, the model focuses on the pre-defined rewards while maintaining high ranking performance. We verify the effectiveness of our approach by integrating SQN and SAC on four state-of-the-art sequential or session-based 
recommendation models.

To summarize, this work makes the following contributions:
\begin{itemize}
    \item We propose self-supervised reinforcement learning for sequential recommendation. Our approach  extends existing recommendation models with a RL layer which aims to introduce reward driven properties to the recommendation.
    \item We propose two frameworks SQN and SAC to co-train the supervised head and the RL head. We integrate four state-of-the-art recommendation models into the proposed frameworks.
    \item We conduct experiments on two real world e-commerce datasets with both clicks and purchases interactions to validate our proposal. Experimental results demonstrate the proposed methods are effective to improve hit ratios, especially when measured against recommending items that eventually got purchased.
\end{itemize}

%% file: preliminaries.tex
\section{Preliminaries}
In this section, we first describe the basic problem of generating next item recommendations from sequential or session-based user data. We introduce reinforcement learning and analyse its limitations for recommendation. Lastly, we provide a literature review on related work.
\subsection{Next Item Recommendation} \label{Next-item-recommendation}
Let $\mathcal{I}$ denote the whole item set, then a user-item interaction sequence can be represented as $x_{1:t}=\left\{x_1,x_2,...,x_{t-1},x_t\right\}$, where $x_i \in \mathcal{I} (0< i \leq t)$ is the index of the interacted item at timestamp $i$. Note that in a real world scenario there may be different kinds of interactions. For instance, in e-commerce use cases, the interactions can be clicks, purchases, add to basket and so on. In video recommendation systems, the interactions can be characterized by the watching time of a video. The goal of next item recommendation is recommending the most relevant item $x_{t+1}$ to the user given the sequence of previous interactions $x_{1:t}$.
\begin{figure*}
    \captionsetup[subfloat]
    {}
    \centering
    \subfloat[Self-supervised training procedure.]{%
    \label{fig:next-item-recommendation-procedure}
    \includegraphics[width=0.24\textwidth]{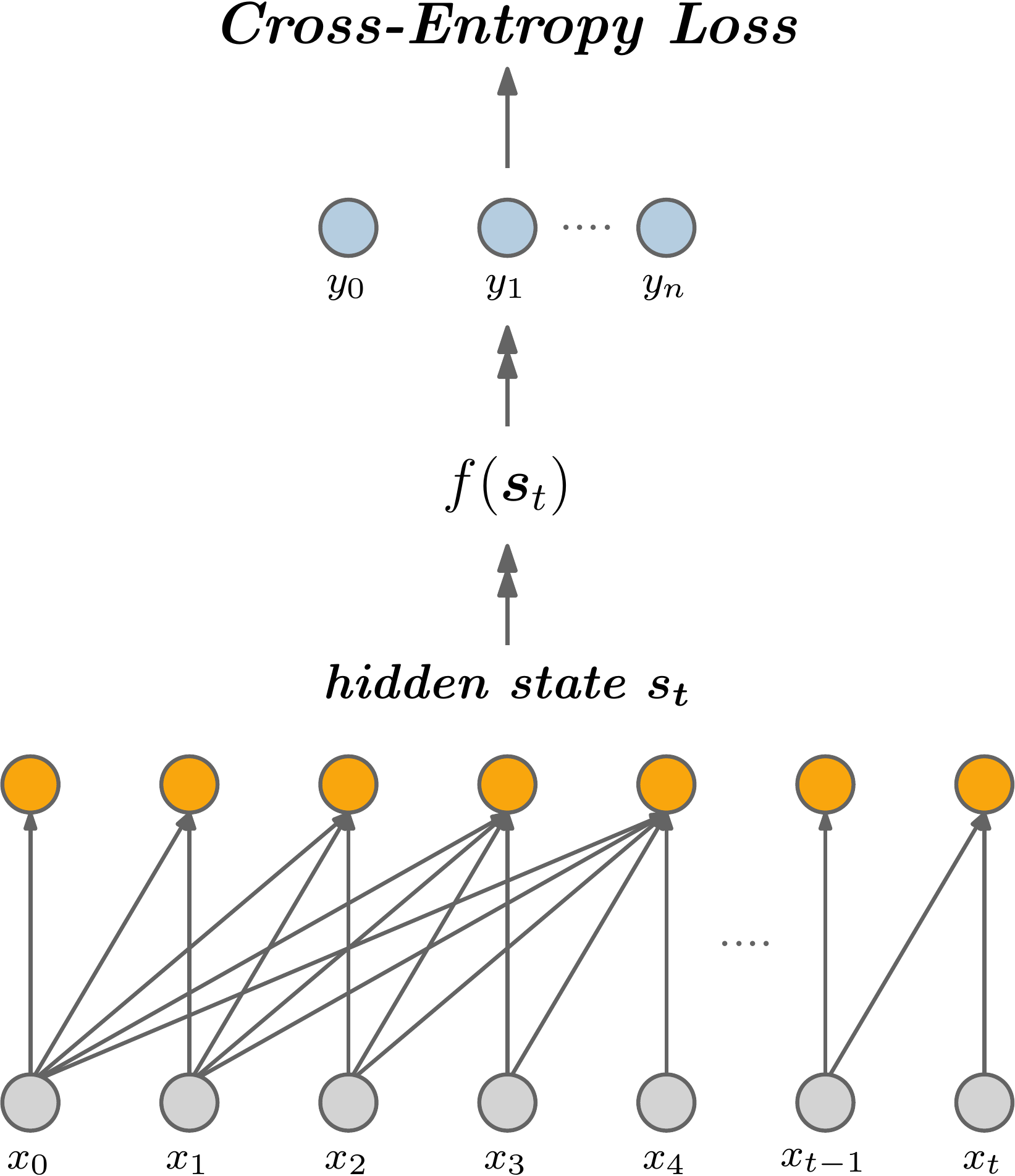}}
    \hspace{0.5cm}
    \subfloat[SQN architecture.]{%
    \label{fig:sqn-architecture}
    \includegraphics[width=0.32\textwidth]{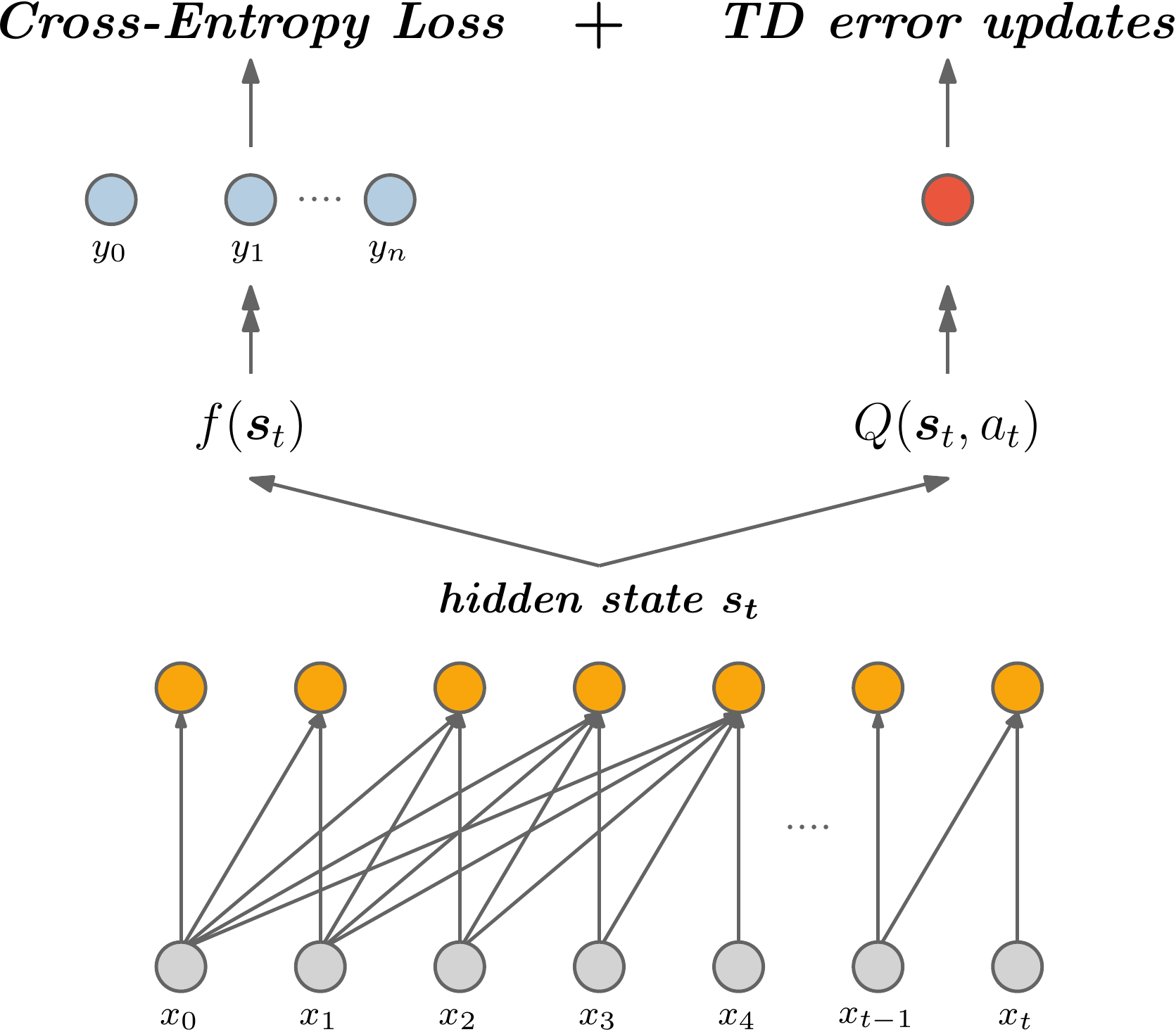}}
    \hspace{0.5cm}
    \subfloat[SAC architecture. CE stands for cross-entropy.]{
    \label{fig:sac-architecture}
    \includegraphics[width=0.32\textwidth]{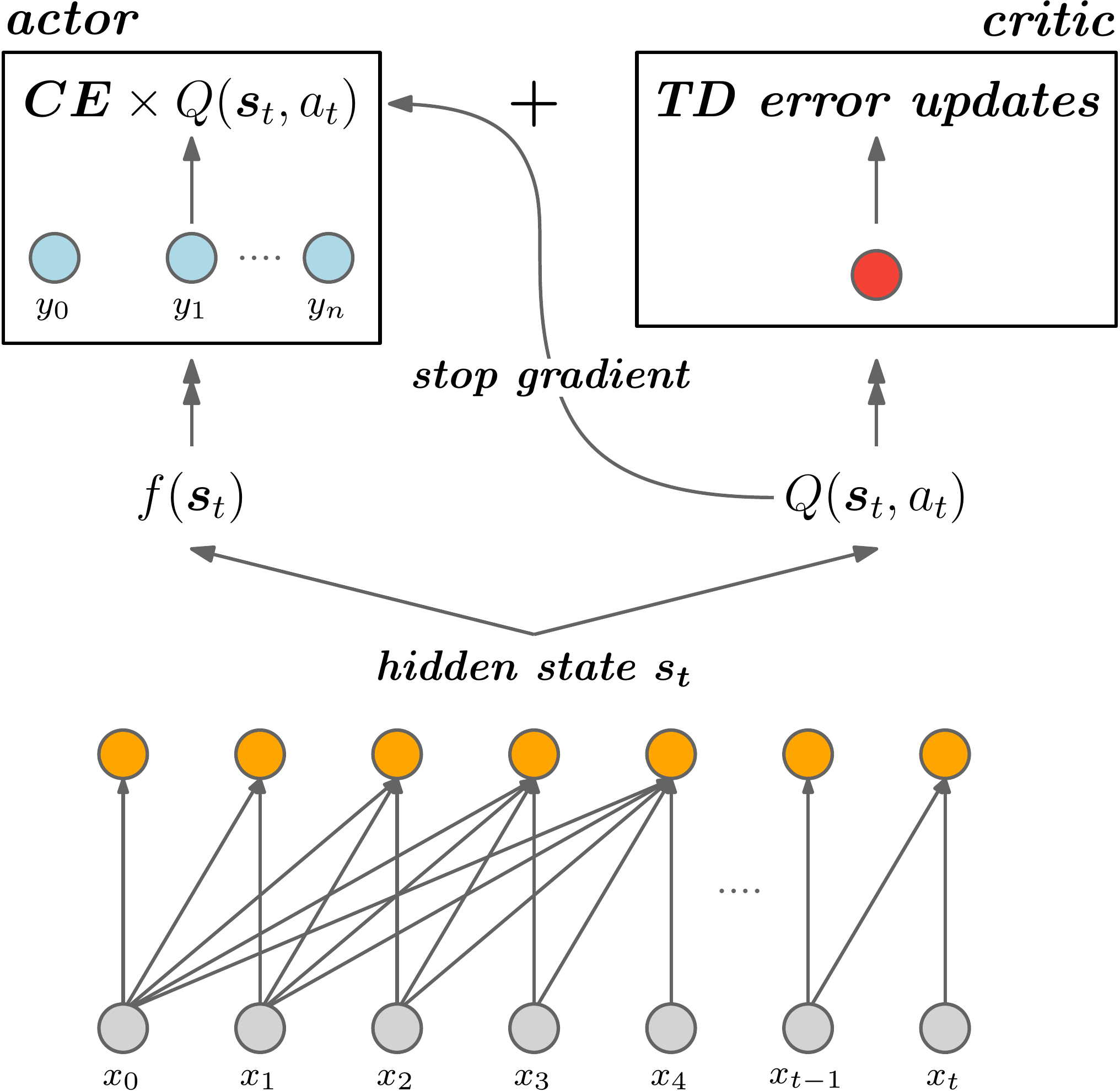}}
    \caption{The self-supervised learning procedure for recommendation and the proposed frameworks.}
\end{figure*}

We can cast this task as a (self-supervised) multi-class classification problem and build a sequential model that generates the classification logits $\mathbf{y}_{t+1}=[y_1,y_2,...y_n] \in \mathbb{R}^n$, where $n$ is the number of candidate items. We can then choose the top-$k$ items from $\mathbf{y}_{t+1}$ as our recommendation list for timestamp $t+1$. A common procedure to train this type of recommender is shown in Figure \ref{fig:next-item-recommendation-procedure}. Typically one can use a generative model $G$ to map the input sequence into a hidden state $\mathbf{s}_t$ as $\mathbf{s}_t=G(x_{1:t})$. This serves as a general encoder function. Plenty of models have been proposed for this task and we will discuss prominent ones in section \ref{related-work}. Based on the obtained hidden state, one can utilize a decoder to map the hidden state to the classification logits as $\mathbf{y}_{t+1}=f(\mathbf{s}_t)$. It is usually defined as a simple fully connected layer or the inner product with candidate item embeddings \cite{nextitnet,gru4rec,caser-rec,SASRec}. Finally, we can train our recommendation model (agents) by optimizing a loss function based on the logits $\mathbf{y}_{t+1}$, such as the cross-entropy loss or the pair-wise ranking loss \cite{rendle2009bpr}.
\subsection{Reinforcement Learning} \label{RL}
In terms of RL, we can formulate the next item recommendation problem as a Markov Decision Process (MDP) \cite{shani2005mdp}, in which the recommendation agent interacts with the environments $\mathcal{E}$ (users) by sequentially recommending items to maximize the long-term cumulative rewards. More precisely, the MDP can be defined by tuples consisting of $(\mathcal{S},\mathcal{A},\mathbf{P},R, \rho_0,\gamma)$ where
\begin{itemize}
    \item $\mathcal{S}$: a continuous state space to describe the user states. This is modeled based on the user (sequential) interactions with the items. The state of the user can be in fact represented by the hidden state of the sequential model discussed in section \ref{Next-item-recommendation}. Hence the state of a user at timestamp $t$ can be represented as $\mathbf{s}_t=G(x_{1:t}) \in \mathcal{S}$ $(t>0)$.
    \item $\mathcal{A}$: a discrete action space which contains candidate items. The action $a$ of the agent is the selected item to be recommended. In off-line data, we can get the action at timestamp $t$ from the user-item interaction (i.e., $a_t=x_{t+1}$). There are also works that focus on generating slate (set)-based recommendations and we will discuss them in section \ref{related-work}.
    \item $\mathbf{P}$: $\mathcal{S} \times \mathcal{A} \times \mathcal{S} \rightarrow \mathbb{R}$ is the state transition probability.
    \item $R$: $\mathcal{S} \times \mathcal{A} \rightarrow \mathbb{R}$ is the reward function, where $r(\mathbf{s},a)$ denotes the immediate reward by taking action $a$ at user state $\mathbf{s}$. The flexible reward scheme is crucial in the utility of RL for recommender systems as it allows for optimizing the recommendation models towards goals that are not captured by conventional loss functions.  For example, in the e-commerce scenario, we can give a larger reward to purchase interactions compared with clicks to build a model that assists the user in his purchase rather than the browsing task. We can also set the reward according to item novelty \cite{diversity} to promote recommendation diversity. For video recommendation, we can set the rewards according to the watching time \cite{googlewsdmoffpolicycorrection}.
    \item $\rho_0$ is the initial state distribution with $\mathbf{s}_0 \sim \rho_0$.
    \item $\gamma$ is the discount factor for future rewards.
\end{itemize}
 RL seeks a target policy $\pi_\theta(a|\mathbf{s})$ which translates the user state $\mathbf{s} \in \mathcal{S}$ into a distribution over actions $a \in \mathcal{A}$, so as to maximize the expected cumulative reward:
\begin{equation}
	\label{cumulative-rewards}
	\max_{\pi_\theta}\mathbb{E}_{\tau\sim\pi_\theta}[R(\tau)]\text{, where }R(\tau)=\sum_{t=0}^{|\tau|}\gamma^{t}r(\mathbf{s}_t,a_t),
\end{equation}
where $\theta \in \mathbb{R}^d$ denotes policy parameters. Note that the expectation is taken over trajectories $\tau=(\mathbf{s}_0,a_0,\mathbf{s}_1,...)$, which are obtained by performing actions according to the target policy: $\mathbf{s}_0 \sim \rho_0$, $a_t \sim \pi_\theta(\cdot|\mathbf{s}_t)$, $s_{t+1} \sim \mathbf{P}(\cdot|\mathbf{s}_t,a_t)$.

Solutions to find the optimal $\theta$ can be categorized into policy gradient-based approaches (e.g., REINFORCE \cite{REINFORCE}) and value-based approaches (e.g., Q-learning \cite{alphogo}). 

Policy-gradient based approaches aim at directly learning the mapping function $\pi_{\theta}$. Using the ``log-trick'' \cite{REINFORCE}, gradients of the expected cumulative rewards with respect to policy parameters $\theta$ can be derived as: 
\begin{equation}
	\label{policy-gradient}
	\mathbb{E}_{\tau\sim\pi_\theta}[R(\tau)\nabla_\theta \log \pi_\theta(\tau)].
\end{equation}


In on-line RL environments, it's easy to estimate the expectation. However, under the recommendation settings, to avoid recommending irrelevant items to users, the agent must be trained using the historical logged data. Even if the RL agent can interact with live users, the actions (recommended items) may be controlled by other recommenders with different policies since many recommendation models might be deployed in a real live recommender system. As a result, what we can estimate from the batched (logged) data is 
\begin{equation}
	\label{gradient-realworld}
	\mathbb{E}_{\tau\sim\beta}[R(\tau)\nabla_\theta \log \pi_\theta(\tau)],
\end{equation}
where $\beta$ denotes the behavior policy that we follow when generating the training data. Obviously, there is distribution discrepancy between the target policy $\pi_\theta$ and the behavior policy $\beta$.  Applying policy-gradient methods for recommendation using this data is thus infeasible. 

\begin{figure}
	\centering
    \includegraphics[width=0.45\textwidth]{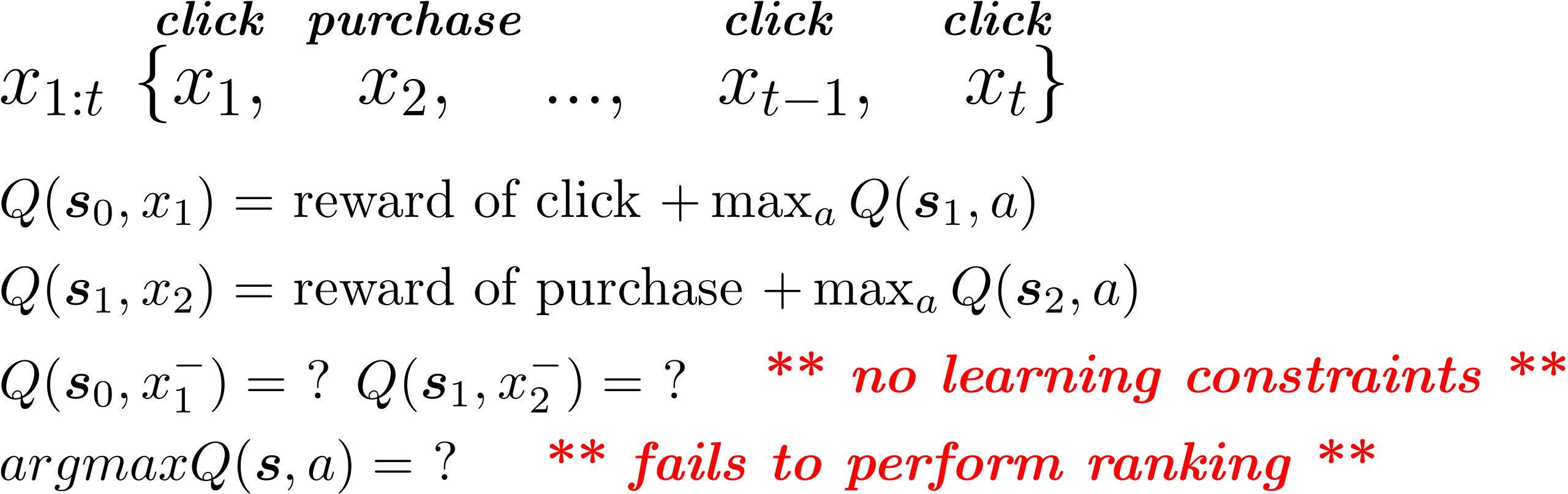}
    \caption{Q-learning fails to learn a proper preference ranking because of data sparsity and the lack of negative feedback. $x_1^-$ and $x_2^-$ are unseen (negative) items for the corresponding timestamp.}
    \label{fig:q-learning-fails}
\end{figure}

Value-based approaches first calculate the Q-values (i.e., $Q(\mathbf{s},a)$, the value of an action $a$ at a given state $\mathbf{s}$) according to the Bellman equation while the action is taken by $a=\text{argmax } Q(\mathbf{s},a)$. The one-step temporal difference (TD) update rule formulates the target $Q(\mathbf{s}_t,a_t)$ as
\begin{equation}
	\label{one-step-q-values}
	Q(\mathbf{s}_t,a_t)=r(\mathbf{s}_{t},a_t)+\gamma \max_{a^{'}}Q(\mathbf{s}_{t+1},a').
\end{equation}
One of the major limitation of implicit feedback data is the lack of negative feedback \cite{rendle2009bpr,yuan2018fbgd}, which means we only know which items the user has interacted with but have no knowledge about the missing transactions. Thus there are limited state-action pairs to learn from and Q-values learned purely on this data would be sub-optimal as shown in Figure \ref{fig:q-learning-fails}.
As a result, taking actions using these Q-values by $a=\text{argmax } Q(\mathbf{s},a)$ would result in poor performance. 
Even though the estimation of $Q(\mathbf{s},a)$ is unbiased due to the greedy selection of Q-learning\footnote{We don't consider the bias introduced by the steps of TD learning. This is not related to our work.}, the distribution of $\mathbf{s}$ in the logged data is biased. So the distribution discrepancy problem of policy gradient-based methods still exists in Q-learning even though the Q-learning algorithm is ``off-policy'' \cite{bcq}.

\subsection{Related Work} \label{related-work}
Early work focusing on sequential recommendation mainly rely on Markov Chain (MC) models \cite{he2016fusingmarkvo,yougonext,rendle2010factorizingmarkov} and factorization-based methods \cite{fm,gmf}. Rendle et. al  \cite{rendle2010factorizingmarkov} introduced to use first-order MC to capture short-term user preferences and combined the MC with matrix factorization (MF) \cite{koren2009mf} to model long-term preferences. Methods with high-order MCs that consider more longer interaction sequences were also proposed in \cite{he2016fusingmarkvo,he2016vista}. Factorization-based methods such as factorization machines (FM) \cite{fm} can utilize the previous items a user has interacted with as context features. The general factorization framework (GFF) \cite{gmf} models a session as the average of the items that the user interacted within that session.

MC-based methods face challenges in modeling complex sequential signals such as skip behaviors in the user-item sequences \cite{caser-rec,nextitnet} while factorization-based methods do not model the order of user-item interactions. As a result, plenty of deep learning-based approaches have been proposed to model the interaction sequences more effectively. \cite{gru4rec} proposed to utilize gated recurrent units (GRU) \cite{GRU} to model the session. \cite{caser-rec} and \cite{nextitnet} utilized convolutional neural networks (CNN) to capture sequential signals. \cite{SASRec} exploited the well-known Transformer \cite{Transformer} in the field of sequential recommendation with promising results. 
Generally speaking, all of these models can serve as the model $G$ described in section \ref{Next-item-recommendation} whose input is a sequence of user-item interactions while the output is a latent representation $\mathbf{s}$ that describes the corresponding user state. 

Attempts to utilize RL for recommendation have also been made. To address the problem of distribution discrepancy under the off-policy settings, \cite{googlewsdmoffpolicycorrection} proposed to utilize propensity scores to perform off-policy correction. However, the estimation of propensity scores has high variances and there is a trade-off between bias and variance, which introduces additional training difficulties. \cite{pairwise-q-learning} proposed to utilize negative sampling along with Q-learning. But their method doesn't address the off-policy problem. Model-based RL approaches \cite{chen2019generativeusermodel,didienvironment,jdkdd19} firstly build a model to simulate the environment in order to avoid any issues with  off-policy training. However, these two-stage approaches heavily depend on the accuracy of the simulator. Moreover, recent work has also been done on providing slate-based recommendations \cite{ie2019slateq,gong2019exactk,googlewsdmoffpolicycorrection,chen2019generativeusermodel} in which actions are considered to be sets (slates) of items to be recommended.  This assumption creates an even larger action space as a slate of items is regarded as one single action. To keep in line with existing self-supervised recommendation models, in this paper we set the action to be recommending the top-$k$ items that are scored by the supervised head. We leave research in this area of set-based recommendation as one of our future work directions. 

Bandit algorithms which share the same reward schema and long-term expectation with RL have also been investigated for recommendation \cite{li2011unbiasedbandit,li2010contextualbandit}. 
Bandit algorithms assume that taking actions does not affect the state \cite{li2010contextualbandit} while in full RL the assumption is that the state is affected by the actions. Generally speaking, recommendations actually have an effect on user behavior \cite{rohde2018recogym} and hence RL is more suitable for modeling the recommendation task.
Another related field is imitation learning where the policy is learned from expert demonstrations \cite{GAIL,ho2016model-free-imitation,torabi2018behavioral-clone}. Our work can be also considered as a form of imitation learning as part of the model is trained to imitate user actions.

%% file: method.tex
\section{Method}
As discussed in section \ref{RL}, directly applying standard RL algorithms to recommender systems data is essentially unfeasible. In this section, we propose to co-train a RL output layer along with the self-supervised head. The reward can be designed according to the specific demands of the recommendation setting. 
We first describe the proposed SQN algorithm and then extend it to SAC. Both algorithms can be easily integrated with existing recommendation models.
\subsection{Self-Supervised Q-learning}
Given an input item sequence $x_{1:t}$ and an existing recommendation model $G$, the self-supervised training loss can be defined as the cross-entropy over the classification distribution:
\begin{equation}
	\label{supervised-loss}
	L_s=-\sum_{i=1}^nY_ilog(p_i), \text{where } p_i=\frac{e^{y_i}}{\sum_{i'=1}^ne^{y_{i'}}}.
\end{equation}
$Y_i$ is an indicator function and $Y_i=1$ if the user interacted with the $i$-th item in the next timestamp. Otherwise, $Y_i=0$. Due to the fact that the recommendation model $G$ has already encoded the input sequence into a latent representation $\mathbf{s}_t$, we can directly utilize $\mathbf{s}_t$ as the state for the RL part without introducing another model. What we need is an additional final layer to calculate the Q-values. A concise solution is stacking a fully-connected layer on the top of $G$:  
\begin{equation}
	\label{q-value-calculation}
	Q(\mathbf{s}_t,a_t)=\delta(\mathbf{s}_t\mathbf{h}_t^T+b)=\delta(G(x_{1:t})\mathbf{h}_t^T+b),
\end{equation}
where $\delta$ denotes the activation function, $\mathbf{h}_t$ and $b$ are trainable parameters of the RL head.

After that, we can define the loss for the RL part based on the one-step TD error:
\begin{equation}
	\label{rl-loss}
	L_q=(r(\mathbf{s}_t,a_t)+\gamma\max_{a'}Q(\mathbf{s}_{t+1},a')-Q(\mathbf{s}_t,a_t))^2
\end{equation}
We jointly train the self-supervised loss and the RL loss on the replay buffer generated from the implicit feedback data:
\begin{equation}
	\label{L-SQN}
	L_{SQN}=L_s+L_q.
\end{equation}
Figure \ref{fig:sqn-architecture} demonstrates the architecture of SQN. 

When generating recommendations, we still return the top-$k$ items from the supervised head. The RL head acts as a regularizer to fine-tune the recommendation model $G$ according to our reward settings. As discussed in section \ref{RL}, the state distribution in the logged data is biased, so generating recommendations using the Q-values is problematic. However, due to the greedy selection of  $Q(\mathbf{s}_{t+1},\cdot)$, the estimation of $Q(\mathbf{s}_{t},a_t)$ itself is unbiased. As a result, by utilizing Q-learning as a regularizer and keeping the self-supervised layer as the source of recommendations we avoid any off-policy correction issues. The lack of negative rewards in Q-learning does also not affect the methods since the RL output layer is trained on positive actions and the supervised cross-entropy loss provides the negative gradient signals which come from the denominator of Eq.(\ref{supervised-loss}). 

To enhance the learning stability, we utilize double Q-learning \cite{double-q-learning} to alternatively train two copies of learnable paramaters. Algorithm 1 describes the training procedure of SQN.
\begin{algorithm}[b]
 \label{alg:SQN}
 \caption{Training procedure of SQN}
 	\begin{algorithmic}[1]
        \renewcommand{\algorithmicrequire}{\textbf{Input:}} 
        \renewcommand{\algorithmicensure}{\textbf{Output:}}
        \Require
        user-item interaction sequence set $\mathcal{X}$, recommendation model $G$, reinforcement head $Q$, supervised head
        \Ensure
        all parameters in the learning space $\Theta$
        \State Initialize all trainable parameters
        \State Create $G'$ and $Q'$ as copies of $G$ and $Q$, respectively
        \Repeat 
            \State Draw a mini-batch of $(x_{1:t},a_t)$ from $\mathcal{X}$, set rewards $r$
            \State $\mathbf{s}_t=G(x_{1:t})$, $\mathbf{s}'_t=G'(x_{1:t})$ 
            \State $\mathbf{s}_{t+1}=G(x_{1:t+1})$, $\mathbf{s}'_{t+1}=G'(x_{1:t+1})$
            \State Generate random variable $z \in (0,1)$ uniformly
            \If {$z\leq0.5$}
                \State $a^*=\text{argmax}_a\text{ }Q(\mathbf{s}_{t+1},a)$
                \State $L_q=(r+\gamma Q'(\mathbf{s}'_{t+1},a^*)-Q(\mathbf{s}_t,a_t))^2$
                \State Calculate $L_s$ and $L_{SQN}$ according to Eq.(\ref{supervised-loss}) and Eq.(\ref{L-SQN})
                \State Perform updates by $\nabla_\Theta L_{SQN}$
            \Else
                \State $a^*=\text{argmax}_a\text{ }Q'(\mathbf{s}'_{t+1},a)$
                \State $L_q=(r+\gamma Q(\mathbf{s}_{t+1},a^*)-Q'(\mathbf{s}'_t,a_t))^2$
                \State Calculate $L_s$ and $L_{SQN}$ according to Eq.(\ref{supervised-loss}) and Eq.(\ref{L-SQN})
                \State Perform updates by $\nabla_\Theta L_{SQN}$
            \EndIf
        \Until converge
        \State return all parameters in $\Theta$
 	\end{algorithmic}
 \end{algorithm}

\subsection{Self-Supervised Actor-Critic}
In the previous subsection, we proposed SQN in which the Q-learning head acts as a regularizer to fine-tune the model in line with the reward schema. The learned Q-values are unbiased and learned from positive user-item interactions (feedback). As a result, these values can be regarded as an unbiased measurement of how the recommended item satisfies our defined rewards. Hence the actions with high Q-values should get increased weight on the self-supervised loss, and vice versa.

We can thus treat the self-supervised head which is used for generating recommendations as a type of ``actor'' and the Q-learning head as the ``critic''. Based on this observation, we can use the Q-values as  weights for the self-supervised loss:
\begin{equation}
	\label{actor-loss}
	L_A=L_s\cdot Q(\mathbf{s}_t,a_t).
\end{equation}
This is similar to what is used in the well-known Actor-Critic (AC) methods \cite{konda2000actor-critic}. However, the actor in AC is based on policy gradient which is on-policy while the ``actor'' in our methods is essentially self-supervised. Moreover, there is only one base model $G$ in SAC while AC has two separate networks for the actor and the critic. 
To enhance stability, we stop the gradient flow and fix the Q-values  when they are used in that case. We then train the actor and critic jointly. Figure \ref{fig:sac-architecture} illustrates the architecture of SAC.
In complex sequential recommendation models (e.g., using the Transformer encoder as $G$ \cite{SASRec}), the learning of Q-values can be unstable \cite{parisotto2019stabilizing}, particularly in the early stages of training. To mitigate these issues, we set a threshold $T$. When the number of update steps is smaller than $T$, we perform normal SQN updates. After that, the Q-values become more stable and we start to use the critic values in the self-supervised layer and perform updates according to the architecture of Figure $\ref{fig:sac-architecture}$. The training procedure of SAC is demonstrated in Algorithm 2.
\begin{algorithm}
 \label{alg:SAC}
 \caption{Training procedure of SAC}
 	\begin{algorithmic}[1]
        \renewcommand{\algorithmicrequire}{\textbf{Input:}} 
        \renewcommand{\algorithmicensure}{\textbf{Output:}}
        \Require
        the interaction sequence set $\mathcal{X}$, recommendation model $G$, reinforcement head $Q$, supervised head, threshold $T$
        \Ensure
        all parameters in the learning space $\Theta$
        \State Initialize all trainable parameters
        \State Create $G'$ and $Q'$ as copies of $G$ and $Q$, $t=0$
        \Repeat 
            \State Draw a mini-batch of $(x_{1:t},a_t)$ from $\mathcal{X}$, set rewards $r$
            \State $\mathbf{s}_t=G(x_{1:t})$, $\mathbf{s}'_t=G'(x_{1:t})$ 
            \State $\mathbf{s}_{t+1}=G(x_{1:t+1})$, $\mathbf{s}'_{t+1}=G'(x_{1:t+1})$
            \State Generate random variable $z \in (0,1)$ uniformly
            \If {$z\leq0.5$}
                \State $a^*=\text{argmax}_a\text{ }Q(\mathbf{s}_{t+1},a)$
                \State $L_q=(r+\gamma Q'(\mathbf{s}'_{t+1},a^*)-Q(\mathbf{s}_t,a_t))^2$
                \If {$t\leq T$}
                    \State Perform updates by $\nabla_\Theta L_{SQN}$
                \Else
                    \State $L_A=L_s\times Q(\mathbf{s}_t,a_t)$, $L_{SAC}=L_A+L_s$
                    \State Perform updates by $\nabla_\Theta L_{SAC}$
                \EndIf
            \Else
                \State $a^*=\text{argmax}_a\text{ }Q'(\mathbf{s}'_{t+1},a)$
                \State $L_q=(r+\gamma Q(\mathbf{s}_{t+1},a^*)-Q'(\mathbf{s}'_t,a_t))^2$
                \If {$t\leq T$}
                    \State Perform updates by $\nabla_\Theta L_{SQN}$
                \Else
                    \State $L_A=L_s\times Q'(\mathbf{s}'_t,a_t)$, $L_{SAC}=L_A+L_s$
                    \State Perform updates by $\nabla_\Theta L_{SAC}$
                \EndIf
            \EndIf
            \State $t=t+1$
        \Until converge
        \State return all parameters in $\Theta$
 	\end{algorithmic}
 \end{algorithm}
\subsection{Discussion}
The proposed training frameworks can be integrated with existing recommendation models, as long as they follow the procedure of Figure \ref{fig:next-item-recommendation-procedure} to generate recommendations. This is the case for most session-based or sequential recommendation models introduced over the last years. In this paper we utilize the cross-entropy loss as the self-supervised loss, while the proposed methods also work for other losses  \cite{rendle2009bpr,hidasi2019topkgains}. The proposed methods are for general purpose recommendation. You can design the reward schema according to your own demands and recommendation objectives. 

Due to the biased state-action distribution in the off-line setting and the lack of sufficient data, directly generating recommendations from RL is difficult. The proposed SQN and SAC frameworks can be seen as attempts to exploit an unbiased RL estimator\footnote{In our case, the estimation of $Q(\mathbf{s},a)$ is unbiased.} to ``reinforce'' existing self-supervised recommendation models.
Another way of looking at the proposed approach is as a form of transfer learning whereby the self-supervised model is used to ``pretrain'' parts of the Q-learning model and vice versa.  

%% file: experiment.tex
\section{Experiments}
In this section, we conduct experiments\footnote{The implementation can be found at \url{https://drive.google.com/open?id=1nLL3_knhj_RbaP_IepBLkwaT6zNIeD5z}} on two real-world sequential recommendation datasets to evaluate the proposed SQN and SAC in the e-commerce scenario. We aim to answer the following research questions:

\textbf{RQ1:} How do the proposed methods perform when integrated with existing recommendation models?
	
\textbf{RQ2:} How does the RL component affect performance, including the reward setting and the discount factor. 

\textbf{RQ3:} What is the performance if we only use Q-leaning for recommendation?

In the following parts, we will describe the experimental settings and answer the above research questions.
\subsection{Experimental Settings}
\subsubsection{Datasets}
We conduct experiments with two public accessible e-commerce datasets: RC15\footnote{\url{https://recsys.acm.org/recsys15/challenge/}} and RetailRocket\footnote{\url{https://www.kaggle.com/retailrocket/ecommerce-dataset}}. 

\textbf{RC15.} This is based on the dataset of RecSys Challange 2015. The dataset is session-based and each session contains a sequence of clicks and purchases. We remove the sessions whose length is smaller than 3 and then sample 200k sessions, resulting into a dataset which contains 1,110,965 clicks and 43,946 purchases over 26702 items. We sort the user-item interactions in one session according to the timestamp.

\textbf{RetailRocket.} This dataset is collected from a real-world e-commerce website. It contains sequential events of viewing and adding to cart. To keep in line with the RC15 dataset, we treat views as clicks and adding to cart as purchases. We remove the items which are interacted less than 3 times and the sequences whose length is smaller than 3. The final dataset contains 1,176,680 clicks and 57,269 purchases over 70852 items.

Table \ref{Datasets} summarizes the statistics of the two datasets.
\begin{table}
    \centering
    \begin{threeparttable}
    \caption{Dataset statistics.}
    \label{Datasets}
    \begin{tabular}{p{1.5cm}p{2.0cm}<{\centering}p{2.0cm}<{\centering}}
    \toprule
    Dataset&RC15 & RetailRocket\cr
    \midrule
    \#sequences&200,000&195,523\cr
    \#items&26,702&70,852\cr
    \#clicks&1,110,965&1,176,680\cr
    \#purchase&43,946&57,269\cr
    \bottomrule
  \end{tabular}
    \end{threeparttable}
\end{table}
\subsubsection{Evaluation protocols}
We adopt cross-validation to evaluate the performance of the proposed methods. The ratio of training, validation, and test set is 8:1:1. We randomly sample 80\% of the sequences as training set. For validation and test sets, the evaluation is done by providing the events of a sequence one-by-one and checking the rank of the item of the next event. The ranking is performed among the whole item set. Each experiment is repeated 5 times, and the average performance is reported.

\begin{table*}
    \centering
    \begin{threeparttable}
    \caption{Top-$k$ recommendation performance comparison of different models ($k=5, 10, 20$) on RC15 dataset. NG is short for NDCG. Boldface denotes the highest score. $*$ denotes the significance $p$-value < 0.01 compared with the corresponding baseline.}
    \label{comparison between different models on RC15}
    \begin{tabular}{p{1.7cm}p{0.85cm}<{\centering}p{0.85cm}<{\centering}p{0.85cm}<{\centering}p{0.85cm}<{\centering}p{0.85cm}<{\centering}p{0.85cm}<{\centering}p{0.85cm}<{\centering}p{0.85cm}<{\centering}p{0.85cm}<{\centering}p{0.85cm}<{\centering}p{0.85cm}<{\centering}p{0.85cm}<{\centering}}
    \toprule
    \multirow{2}{*}{Models}&\multicolumn{6}{c}{purchase}&\multicolumn{6}{c}{click}\cr
    \cmidrule(lr){2-7} \cmidrule(lr){8-13}
    &HR@5&NG@5&HR@10&NG@10&HR@20&NG@20&HR@5&NG@5&HR@10&NG@10&HR@20&NG@20\cr
    \midrule
    GRU    &0.3994&0.2824 &0.5183&0.3204 &0.6067&0.3429 &0.2876&0.1982 &0.3793&0.2279 &0.4581&0.2478\cr
    GRU-SQN&$0.4228^*$&$0.3016^*$&$0.5333^*$&$0.3376^*$&$0.6233^*$&$0.3605^*$&$\mathbf{0.3020^*}$ &$\mathbf{0.2093^*}$ &$\textbf{0.3946}^*$&$\textbf{0.2394}^*$&$\textbf{0.4741}^*$&$\textbf{0.2587}^*$\cr
    GRU-SAC&$\textbf{0.4394}^*$&$\textbf{0.3154}^*$&$\textbf{0.5525}^*$&$\textbf{0.3521}^*$&$\textbf{0.6378}^*$&$\textbf{0.3739}^*$&0.2863 &0.1985 &0.3764 &0.2277 &0.4541&0.2474\\ \hline
    Caser& 0.4475&0.3211 &0.5559&0.3565&0.6393&0.3775&0.2728 &0.1896&0.3593&0.2177&0.4371&0.2372\cr
    Caser-SQN&$0.4553^*$&$0.3302^*$&$0.5637^*$&$0.3653^*$&$0.6417^*$&$0.3862^*$&\textbf{0.2742}&\textbf{0.1909} &\textbf{0.3613}&\textbf{0.2192}&\textbf{0.4381}&\textbf{0.2386}\cr
    Caser-SAC&$\textbf{0.4866}^*$&$\textbf{0.3527}^*$&$\textbf{0.5914}^*$&$\textbf{0.3868}^*$&$\textbf{0.6689}^*$&$\textbf{0.4065}^*$&0.2726&0.1894&0.3580&0.2171&0.4340&0.2362\\ \hline
    NItNet&0.3632&0.2547&0.4716&0.2900&0.5558&0.3114&0.2950&0.2030&0.3885&0.2332&0.4684&0.2535\cr
    NItNet-SQN&$0.3845^*$&$0.2736^*$&$0.4945^*$&$0.3094^*$&$\textbf{0.5766}^*$&$0.3302^*$&$\textbf{0.3091}^*$ &$\textbf{0.2137}^*$ &$\textbf{0.4037}^*$ &$\textbf{0.2442}^*$&$\textbf{0.4835}^*$&$\textbf{0.2645}^*$\cr
    NItNet-SAC&$\textbf{0.3914}^*$&$\textbf{0.2813}^*$&$\textbf{0.4964}^*$&$\textbf{0.3155}^*$&$0.5763^*$&$\textbf{0.3357}^*$&$0.2977^*$&$0.2055^*$&0.3906&$0.2357^*$&0.4693&$0.2557^*$\\ \hline
    SASRec& 0.4228&0.2938 &0.5418&0.3326&0.6329&0.3558&0.3187&0.2200&0.4164&0.2515&0.4974&0.2720\cr
    SASRec-SQN&0.4336&$0.3067^*$&0.5505&$0.3435^*$& $0.6442^*$&$0.3674^*$ &$\textbf{0.3272}^*$&$\textbf{0.2263}^*$&$\textbf{0.4255}^*$&$\textbf{0.2580}^*$&$\textbf{0.5066}^*$&$\textbf{0.2786}^*$\cr
    SASRec-SAC&$\textbf{0.4540}^*$&$\textbf{0.3246}^*$&$\textbf{0.5701}^*$&$\textbf{0.3623}^*$&$\textbf{0.6576}^*$& $\textbf{0.3846}^*$&0.3130&0.2161&0.4114&0.2480&0.4945&0.2691\\
    \bottomrule
    \end{tabular}
    \end{threeparttable}
\end{table*}

\begin{table*}
    \centering
    \begin{threeparttable}
    \caption{Top-$k$ recommendation performance comparison of different models ($k=5, 10, 20$) on RetailRocket. NG is short for NDCG. Boldface denotes the highest score. $*$ denotes the significance $p$-value < 0.01 compared with the corresponding baseline.}
    \label{comparison between different models on RetailRocket}
    \begin{tabular}{p{1.7cm}p{0.85cm}<{\centering}p{0.85cm}<{\centering}p{0.85cm}<{\centering}p{0.85cm}<{\centering}p{0.85cm}<{\centering}p{0.85cm}<{\centering}p{0.85cm}<{\centering}p{0.85cm}<{\centering}p{0.85cm}<{\centering}p{0.85cm}<{\centering}p{0.85cm}<{\centering}p{0.85cm}<{\centering}}
    \toprule
    \multirow{2}{*}{Models}&\multicolumn{6}{c}{purchase}&\multicolumn{6}{c}{click}\cr
    \cmidrule(lr){2-7} \cmidrule(lr){8-13}
    &HR@5&NG@5&HR@10&NG@10&HR@20&NG@20&HR@5&NG@5&HR@10&NG@10&HR@20&NG@20\cr
    \midrule
    GRU    &0.4608 &0.3834&0.5107 &0.3995&0.5564&0.4111&0.2233&0.1735&0.2673&0.1878&0.3082&0.1981\cr
    GRU-SQN&$\textbf{0.5069}^*$&$0.4130^*$&$\textbf{0.5589}^*$&$0.4289^*$&$\textbf{0.5946}^*$&$0.4392^*$&$\textbf{0.2487}^*$ &$\textbf{0.1939}^*$ &$\textbf{0.2967}^*$&$\textbf{0.2094}^*$&$\textbf{0.3406}^*$&$\textbf{0.2205}^*$\cr
    GRU-SAC&$0.4942^*$&$\textbf{0.4179}^*$&$0.5464^*$&$\textbf{0.4341}^*$&$0.5870^*$&$\textbf{0.4428}^*$&$0.2451^*$&$0.1924^*$&$0.2930^*$ &$0.2074^*$&$0.3371^*$&$0.2186^*$\\ \hline
    Caser& 0.3491&0.2935 &0.3857&0.3053&0.4198&0.3141&0.1966&0.1566&0.2302&0.1675&0.2628&0.1758\cr
    Caser-SQN&$0.3674^*$&$0.3089^*$&$0.4050^*$&$0.3210^*$&$0.4409^*$&$0.3301^*$&$0.2089^*$&$0.1661^*$ &$0.2454^*$&$0.1778^*$&$0.2803^*$&$0.1867^*$\cr
    Caser-SAC&$\textbf{0.3871}^*$&$\textbf{0.3234}^*$&$\textbf{0.4336}^*$&$\textbf{0.3386}^*$&$\textbf{0.4763}^*$&$\textbf{0.3494}^*$&$\textbf{0.2206}^*$&$\textbf{0.1732}^*$&$\textbf{0.2617}^*$&$\textbf{0.1865}^*$&$\textbf{0.2999}^*$&$\textbf{0.1961}^*$\\ \hline
    NItNet&0.5630&0.4630&0.6127&0.4792&0.6477&0.4881&0.2495&0.1906&0.2990&0.2067&0.3419&0.2175\cr
    NItNet-SQN&$\textbf{0.5895}^*$&$0.4860^*$&$\textbf{0.6403}^*$&$0.5026^*$&$\textbf{0.6766}^*$&$0.5118^*$&$\textbf{0.2610}^*$ &$\textbf{0.1982}^*$ &$\textbf{0.3129}^*$ &$\textbf{0.2150}^*$&$\textbf{0.3586}^*$&$\textbf{0.2266}^*$\cr
    NItNet-SAC&$\textbf{0.5895}^*$&$\textbf{0.4985}^*$&$0.6358^*$&$\textbf{0.5162}^*$&$0.6657^*$&$\textbf{0.5243}^*$&$0.2529^*$&$0.1964^*$&$0.3010^*$&$0.2119^*$&$0.3458^*$&$0.2233^*$\\ \hline
    SASRec&0.5267&0.4298&0.5916&0.4510&0.6341&0.4618&0.2541&0.1931&0.3085&0.2107&0.3570&0.2230\cr
    SASRec-SQN&$\textbf{0.5681}^*$&$0.4617^*$&$\textbf{0.6203}^*$&$0.4806^*$&$\textbf{0.6619}^*$&$0.4914^*$&$\textbf{0.2761}^*$&$\textbf{0.2104}^*$&$\textbf{0.3302}^*$&$\textbf{0.2279}^*$&$\textbf{0.3803}^*$&$\textbf{0.2406}^*$\cr
    SASRec-SAC&$0.5623^*$&$\textbf{0.4679}^*$&$0.6127^*$&$\textbf{0.4844}^*$&$0.6505^*$& $\textbf{0.4940}^*$&$0.2670^*$&$0.2056^*$&$0.3208^*$&$0.2230^*$&$0.3701^*$&$0.2355^*$\\
    \bottomrule
    \end{tabular}
    \end{threeparttable}
\end{table*}

The recommendation quality is measured with two metrics: hit ration (HR) and normalized discounted cumulative gain (NDCG). HR@$k$ is a recall-based metric, measuring whether the ground-truth item is in the top-$k$ positions of the recommendation list. We can define HR for  clicks as:
\begin{equation}
	\label{HR_click}
	\text{HR(click)}=\frac{\#\text{hits among clicks}}{\#\text{clicks}}.
\end{equation}
HR(purchase) is defined similarly with HR(click) by replacing the clicks with purchases. NDCG is a rank sensitive metric which assign higher scores to top positions in the recommendation list \cite{NDCG}.

\subsubsection{Baselines}
We integrated the proposed SQN and SAC with four state-of-the-art (generative) sequential recommendation models:
\begin{itemize}
    \item GRU \cite{gru4rec}: This method utilizes a GRU to model the input sequences. The final hidden state of the GRU is treated as the latent representation for the input sequence.
	\item Caser \cite{caser-rec}: This is a recently proposed CNN-based method which captures sequential signals by applying convolution operations on the embedding matrix of previous interacted items.
	\item NItNet \cite{nextitnet}: This method utilizes dilated CNN to enlarge the receptive field and residual connection to increase the network depth, achieving good performance with high efficiency.
	\item SASRec \cite{SASRec}: This baseline is motivated from self-attention and uses the Transformer \cite{Transformer} architecture. The output of the Transformer encoder is treated as the latent representation for the input sequence. 
\end{itemize}
\begin{figure*}
    \captionsetup[subfloat]
    {}
    \centering
    \subfloat[SQN for purchase]{
    \label{ratio-sqn-purchase-rc}
    \includegraphics[width=0.24\textwidth]{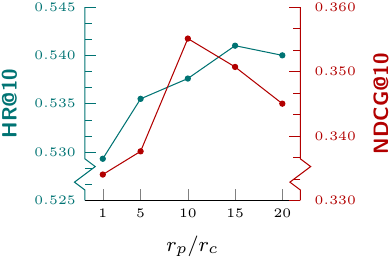}}
    \hspace{0.5mm}
    \subfloat[SQN for click]{%
    \label{ratio-sqn-click-rc}
    \includegraphics[width=0.24\textwidth]{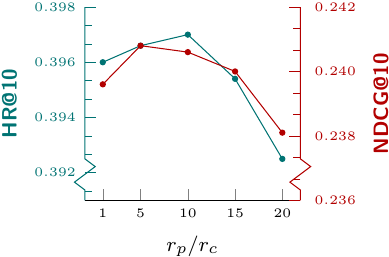}}
    \hspace{0.5mm}
    \subfloat[SAC for purchase]{%
    \label{ratio-sac-purchase-rc}
    \includegraphics[width=0.24\textwidth]{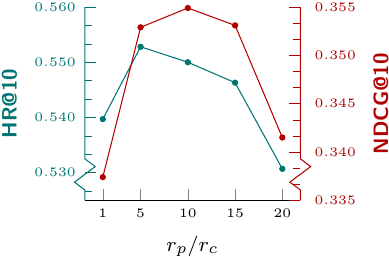}}
    \hspace{0.5mm}
    \subfloat[SAC for click]{%
    \label{ratio-sac-click-rc}
    \includegraphics[width=0.24\textwidth]{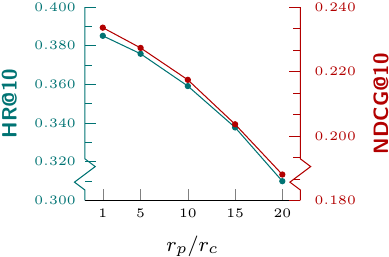}}
    \caption{Effect of reward settings on RC15}
    \label{ratio-RC}
\end{figure*}

\begin{figure*}
    \captionsetup[subfloat]
    {}
    \centering
    \subfloat[SQN for purchase]{
    \label{ratio-sqn-purchase-retail}
    \includegraphics[width=0.24\textwidth]{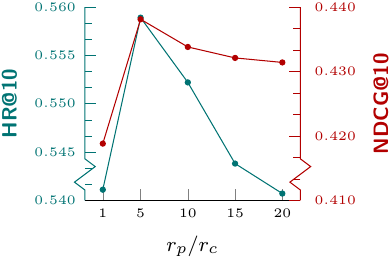}}
    \hspace{0.5mm}
    \subfloat[SQN for click]{%
    \label{ratio-sqn-click-retail}
    \includegraphics[width=0.24\textwidth]{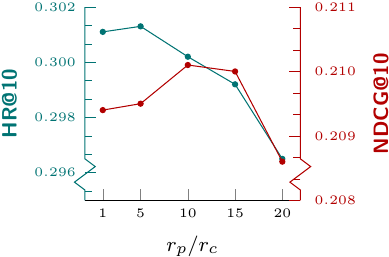}}
    \hspace{0.5mm}
    \subfloat[SAC for purchase]{%
    \label{ratio-sac-purchase-retail}
    \includegraphics[width=0.24\textwidth]{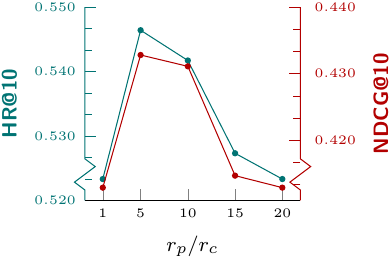}}
    \hspace{0.5mm}
    \subfloat[SAC for click]{%
    \label{ratio-sac-click-retail}
    \includegraphics[width=0.24\textwidth]{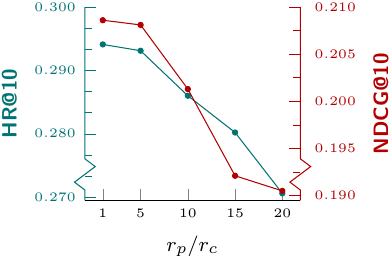}}
    \caption{Effect of reward settings on RetailRocket}
    \label{ratio-Kaggle}
\end{figure*}

\subsubsection{Parameter settings}
For both of the datasets, the input sequences are composed of the last 10 items before the target timestamp. 
If the sequence length is less than 10, we complement the sequence with a padding item.
We train all models with the Adam optimizer \cite{kingma2014adam}. The mini-batch size is set as 256. The learning rate is set as 0.01 for RC15 and 0.005 for RetailRocket. We evaluate on the validation set every 2000 batches of updates. For a fair comparison, the item embedding size is set as 64 for all models. For GRU4Rec, the size of the hidden state is set as 64. For Caser, we use 1 vertical convolution filter and 16 horizontal filters whose heights are set from \{2,3,4\}. The drop-out ratio is set as 0.1. For NextItNet, we utilize the code published by its authors and keep the settings unchanged. For SASRec, the number of heads in self-attention is set as 1 according to its original paper \cite{SASRec}.  For the proposed SQN and SAC, if not mentioned otherwise, the discount factor $\gamma$ is set as 0.5 while the ratio between the click reward ($r_c$) and the purchase reward ($r_p$) is set as $r_p/r_c=5$.

\subsection{Performance Comparison (RQ1)}
Table \ref{comparison between different models on RC15} and Table \ref{comparison between different models on RetailRocket} show the performance of top-$k$ recommendation on RC15 and RetailRocket, respectively.

We observe that on the RC15 dataset, the proposed SQN method achieves consistently better performance than the corresponding baseline when predicting both clicks and purchases. SQN introduces a Q-learning head which aims to model the long-term cumulative reward. The additional learning signal from this head improves both clicks and purchase recommendation performance because the models are now trained to select actions which are optimized not only for the current state but also for future interactions.
We can see that on this dataset, the best performance when predicting purchase interactions is achieved by SAC. Since the learned Q-values are used as weights for the supervised loss function, the model is ''reinforced'' to focus more on purchases. As a result, the SAC method achieves significant better results when recommending purchases. We can assume that the strong but sparse signal that comes with a purchase is better utilized by SAC. 

On the RetailRocket dataset, we can see that both SQN and SAC achieve consistent better performance than the corresponding baseline in terms of predicting both clicks and purchases. This further verifies the effectiveness of the proposed methods. Besides, we can also see that even though SQN sometimes achieves the highest HR(purchase), SAC always achieves the best performance with respect to the NDCG of purchase. This demonstrates that the proposed SAC is actually more likely to push the items which may lead to a purchase to the top positions of the recommendation list. 

To conclude, it's obvious that the proposed SQN and SAC achieve consistent improvement with respect to the selected baselines. This demonstrates the effectiveness and the generalization ability of our methods.

\subsection{RL Investigation(RQ2)}
\subsubsection{Effect of reward settings.}
In this part, we conduct experiments to investigate how the reward setting of RL affects the model performance. Figure \ref{ratio-RC} and Figure \ref{ratio-Kaggle} show the results of HR@10 and NDCG@10 when changing the ratio between $r_p$ and $r_c$ (i.e., $r_p/r_c$) on RC15 and RetailRocket, respectively. We show the performance when choosing GRU as the base model. Results when utilizing other baseline models show similar trends and are omitted due to space limitation.

We can see from Figure \ref{ratio-sqn-purchase-rc} and Figure \ref{ratio-sqn-purchase-retail} that the performance of SQN when predicting purchase interactions start to improve when $r_p/r_c$ increases from 1. It shows that when we assign a higher reward to purchases, the introduced RL head successfully drives the model to focus on the desired rewards. Performance start to decrease after reaching higher ratios. 
The reason may be that high reward differences might cause instability in the TD updates and thus affects the model performance.
Figure \ref{ratio-sac-purchase-rc} and Figure \ref{ratio-sac-purchase-retail} shows that the performance of SAC when predicting purchase behaviours also improves at the beginning and then drops with the increase of $r_p/r_c$. It's similar with SQN.

For click recommendation, we can see from Figure \ref{ratio-sqn-click-rc} and Figure \ref{ratio-sqn-click-retail} that the performance of SQN is actually stable at the beginning (even increases a little) and then starts to decrease. There are two factors for this performance drop. The first is the instability of RL as discussed before.
The second is that too much reward discrepancy might diminish the relative importance of clicks which constitute the vast majority of the datapoints. This also helps to explain the performance drop of SAC as shown in Figure \ref{ratio-sac-click-rc} and Figure \ref{ratio-sac-click-retail}.

\begin{figure}
\setlength{\abovecaptionskip}{0.2cm}
    \captionsetup[subfloat]
    {}
    \centering
    \subfloat[SQN for purchase]{
    \label{discount-sqn-purchase}
    \includegraphics[width=0.23\textwidth]{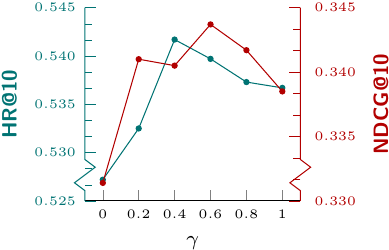}}
    \hspace{0.5mm}
    \subfloat[SQN for click]{%
    \label{discount-sqn-click}
    \includegraphics[width=0.23\textwidth]{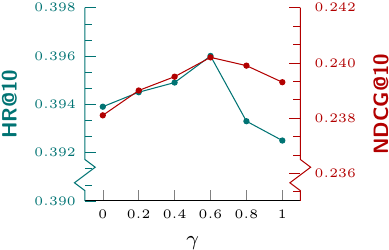}}
    \caption{SQN with different discount factors}
    \label{discount-SQN}
\end{figure}
\begin{figure}
\setlength{\abovecaptionskip}{0.2cm}
    \captionsetup[subfloat]
    {}
    \centering
    \subfloat[SAC for purchase]{
    \label{discount-sac-purchase}
    \includegraphics[width=0.23\textwidth]{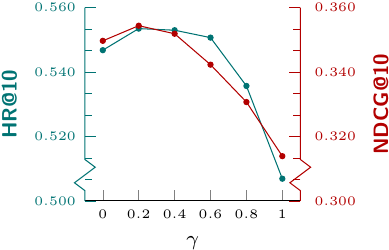}}
    \hspace{0.5mm}
    \subfloat[SAC for click]{%
    \label{discount-sac-click}
    \includegraphics[width=0.23\textwidth]{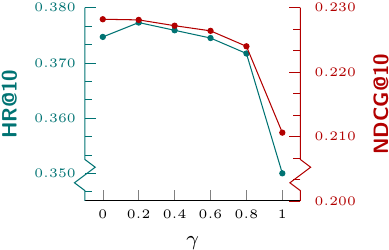}}
    \caption{SAC with different discount factors}
    \label{discount-SAC}
\end{figure}
In addition, observing the performance of SQN and SAC when $r_p/r_c=1$, we can find that they still perform better than the basic GRU. For example, when predicting purchases on the RC15 dataset, the HR@10 of SAC is around 0.54 according to Figure \ref{ratio-sac-purchase-rc} while the basic GRU method only achieves 0.5183 according to Table \ref{comparison between different models on RC15}. This means that even if we don't distinguish between clicks and purchases, the proposed SQN and SAC still works better than the basic model.  The reason is that the introduced RL head successfully adds additional learning signals for long-term rewards.
\subsubsection{Effect of the discount factor}
In this part, we show how the discount factor affects the recommendation performance. Figure \ref{discount-SQN} and Figure \ref{discount-SAC} illustrates the HR@10 and NDCG@10 of SQN and SAC with different discount factors on the RC15 dataset. We choose GRU as the base recommendation model. The results on RetailRocket show similar trends and are omitted here. We can see that the performance of SQN and SAC improves when the discount factor $\gamma$ increases from 0. $\gamma=0$ means that the model doesn't consider long-term reward and only focuses on immediate feedback. This observation leads to the conclusion that taking long-term rewards into account does improve the overall HR and NDCG on both click and purchase recommendations. However, we can also see the performance decreases when the discount factor is too large. Compared with the game control domain in which there maybe thousands of steps in one MDP, the user interaction sequence is much shorter. The average sequence length of the two datasets is only 6. As a result, although $\gamma=0.95 \text{ or } 0.99$ is a common setting for game control, a smaller discount factor should be applied under the recommendation settings.
\subsection{Q-learning for Recommendation (RQ3)}
\begin{figure}
    \captionsetup[subfloat]
    {}
    \centering
    \subfloat[Purchase predictions]{
    \label{q-learning-rc-purchase-hr}
    \includegraphics[width=0.23\textwidth]{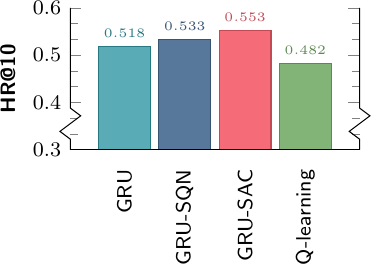}}
    \hspace{0.5mm}
    \subfloat[Click predictions]{%
    \label{q-learning-rc-click-hr}
    \includegraphics[width=0.23\textwidth]{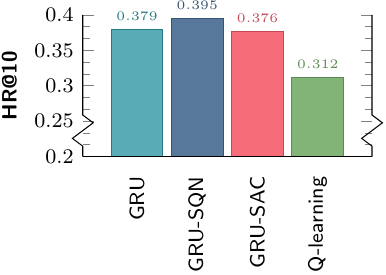}}
    \caption{Comparison of HR when only using Q-learning for recommendations.}
    \label{q-learning-rc-hr}
\end{figure}

\begin{figure}
    \captionsetup[subfloat]
    {}
    \centering
    \subfloat[Purchase predictions]{
    \label{q-learning-rc-purchase-ndcg}
    \includegraphics[width=0.23\textwidth]{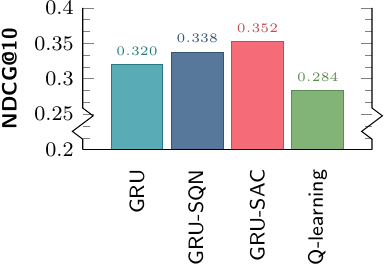}}
    \hspace{0.5mm}
    \subfloat[Click predictions]{%
    \label{q-learning-rc-click-ndcg}
    \includegraphics[width=0.23\textwidth]{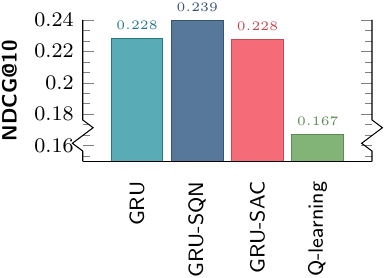}}
    \caption{Comparison of NDCG when only using Q-learning for recommendations.}
    \label{q-learning-rc-ndcg}
    \vspace{-0.2cm}
\end{figure}
We also conduct experiments to examine the performance if we generate recommendations only using Q-learning. To make Q-learning more effective when perform ranking, we explicitly introduce uniformly sampled unseen items to provide negative rewards \cite{rendle2009bpr,pairwise-q-learning}. Figure \ref{q-learning-rc-hr} and Figure \ref{q-learning-rc-ndcg} show the results in terms of HR@10 and NDCG@10 on the RC15 dataset, respectively. The base model is GRU. We can see that the performance of Q-learning is even worse than the basic GRU method. As discussed in section \ref{RL}, directly utilizing Q-learning for recommendation is problematic and off-policy correction needs to be considered in that situation. However, the estimation of Q-values based on the given state is unbiased and exploiting Q-learning as a regularizer or critic doesn't suffer from the above problem. Hence the proposed SQN and SAC achieve better performance.

%% file: conclusion.tex
\section{Conclusion and Future Work}
We propose self-supervised reinforcement learning for recommender systems. We first formalize the next item recommendation task and then analysis the difficulties when exploiting RL for this task. The first is the pure off-policy setting which means the recommender agent must be trained from logged data without interactions between the environment. The second is the lack of negative rewards. 
To address these problems, we propose to augment the existing recommendation model with another RL head. This head acts as a regularizer to introduce our specific desires into the recommendation. The motivation is to utilize the unbiased estimator of RL to fine-tune the recommendation model according to our own reward settings. Based on that, we propose SQN and SAC to perform joint training of the supervised head and the RL head. To verify the effectiveness of our methods, we integrate them with four state-of-the-art recommendation models and conduct experiments on two real-world e-commerce datasets. Experimental results demonstrate that the proposed SQN and SAC are effective to improve the hit ratio, especially when predicting the real purchase interactions.
Future work includes online tests and more experiments on other use cases, such as recommendation diversity promotion, improving watching time for video recommendation and so on. Besides, we are also trying to extend the framework for slate-based recommendation in which the action is recommending a set of items.